\documentclass{IEEEtran}

\usepackage{amsmath,amsfonts}
\usepackage{algorithm,algpseudocode}
\usepackage{todonotes}
\usepackage{amsmath}
\usepackage{color}
\usepackage{cite}
\usepackage{cleveref}
\usepackage{booktabs}
\usepackage{siunitx}
\usepackage[hyphens]{url}

\newcommand{\intrid}{\mathrm{i}}
\newcommand{\ownid}{\mathrm{o}}
\newcommand{\intr}[1]{\ensuremath{#1^{(\intrid)}}}
\newcommand{\own}[1]{\ensuremath{#1^{(\ownid)}}} 
\newcommand{\either}[1]{\ensuremath{#1^{(\cdot)}}}

\newcommand{\is}{\ensuremath{\intr{s}}}
\newcommand{\rs}{\ensuremath{\own{s}}}

\newcommand{\psires}{\ensuremath{\own{\psi}_\mathrm{resolution}}}
\newcommand{\psigoal}{\own{\psi}_\mathrm{goal}}

\newcommand{\sterm}{\ensuremath{s_{\mathrm{term}}}}

\newcommand{\psicand}{\own{\psi}_{\mathrm{cand}}}
\newcommand{\dgoal}{\ensuremath{D_\mathrm{goal}}}
\newcommand{\dnmac}{\ensuremath{D_\mathrm{NMAC}}}

\newcommand{\dev}{\ensuremath{\mathrm{dev }}}

\newcommand{\argmax}[1]{\underset{#1}{\operatornamewithlimits{argmax}}}
\newcommand{\argmin}[1]{\underset{#1}{\operatornamewithlimits{argmin}}}

\newcommand{\real}{{\mathbb{R}}}
\newcommand{\reals}{\real}
\renewcommand{\natural}{{\mathbb{N}}}
\newcommand{\naturals}{\natural}

\algtext*{EndFunction}
\algtext*{EndIf}

\newcommand{\captioncheat}{}

\title{Optimized and Trusted Collision Avoidance for \\ Unmanned Aerial Vehicles using Approximate Dynamic Programming (Technical Report)}

\author{Zachary N. Sunberg, Mykel J. Kochenderfer, and Marco Pavone
\thanks{Zachary Sunberg, Mykel Kochenderfer, and Marco Pavone are with the Department of Aeronautics \& Astronautics, Stanford University, Stanford, CA 94305 {\tt\footnotesize \{zsunberg, mykel, pavone\}@stanford.edu}}
}
\begin{document}
\maketitle

\begin{abstract}
Safely integrating unmanned aerial vehicles into civil airspace is contingent upon development of a trustworthy collision avoidance system. This paper proposes an approach whereby a parameterized resolution logic that is considered \emph{trusted} for a given range of its parameters is adaptively tuned online. Specifically, to address the potential conservatism of the resolution logic with static parameters, we present a  dynamic programming approach for adapting the parameters dynamically based on the encounter state. We compute the adaptation policy offline using a simulation-based approximate dynamic programming method that accommodates the high dimensionality of the problem. Numerical experiments show that this approach improves safety and operational performance compared to the baseline resolution logic, while retaining trustworthiness.
\end{abstract}

\section{Introduction}\label{sec:intro}

%Currently, unmanned aerial vehicles (UAVs) are operated under close human supervision. 
As unmanned aerial vehicles (UAVs) move toward full autonomy, it is vital that they be capable of effectively responding to anomalous events, such as the intrusion of another aircraft into the vehicle's flight path. Minimizing collision risk for aircraft in general, and UAVs in particular, is challenging for a number of reasons. First, avoiding collision requires planning in a way that accounts for the large degree of uncertainty in the future paths of the aircraft. Second, the planning process must balance the competing goals of ensuring safety and avoiding disruption of normal operations. Many approaches have been proposed to address these challenges \cite{JKK-LCY:00,RB-CF-HE:09,GH-RB-JM:11,HH-JJ-CM:10,AN-CM-GD:12,MJK-JPC:11,HB-DH-MJK-WSL:12,JEH-MJK-WAO:13,EJR:14}.

At present, there are two fundamentally different approaches to designing a conflict resolution system. %The first approach focuses on making such a system as easy as possible to certify by a government regulator, vehicle manufacturer, or vehicle operator by making the algorithm as easy as possible for a human to understand. 
% The first approach focuses on making the resolution algorithms as easy as possible for a human to understand to facilitate the confidence of government  in the system government regulators and vehicle operators. 
The first approach focuses on inspiring confidence and trust in the system by making it as simple as possible for regulators and vehicle operators to understand by using hand-specified rules. Several algorithms that fit this paradigm have been proposed, and research toward formally verifying their safety-critical properties is underway \cite{RB-CF-HE:09,GH-RB-JM:11,HH-JJ-CM:10,AN-CM-GD:12}.  In this paper, we will  refer to such algorithms as \emph{trusted resolution logics} (TRLs). TRLs typically have a number of parameters that determine how conservatively the system behaves, a feature that will be exploited later in this paper.
% A TRL typically consists of relatively simple hand-specified logic, and has a number of  parameters that determine how conservatively it behaves.

The second approach focuses on optimizing performance. This entails the offline or online  computation of a ``best" response action. Dynamic programming is widely used for this task \cite{MJK-JPC:11,HB-DH-MJK-WSL:12,JEH-MJK-WAO:13}. Conflict resolution systems designed using this approach will be referred to as \emph{directly optimized} systems. Unfortunately, even if a conflict resolution system performs well in simulation, government regulators and vehicle operators will often (and sometimes rightly) be wary of trusting its safety due to perceived complexity and unpredictability. Even in the best case, such a system would require expensive and time-consuming development of tools for validation as was the case for the recently developed replacement for the traffic alert and collision avoidance system (TCAS) \cite{JEH-MJK-WAO:13}. 

In this paper, we propose a conflict resolution strategy that combines the strengths of these two approaches. The key idea is to use dynamic programming to find a policy that actively adjusts the parameters of a TRL to improve performance. If this TRL is trusted and certified for a range of parameter values, then the optimized version of the TRL should also be trusted and more easily certifiable.
% Thus, the system yielded by the new approach retains the trustedness of the TRL while also improving performance.
% The motivation behind this approach is that as long as the original TRL is trusted and certified for the range of parameters that a policy can select from, the difficulty of certification is not increased beyond what is required for TRL's certification. 

\begin{figure}[tb]
    \centering
    \includegraphics[width=0.6\columnwidth]{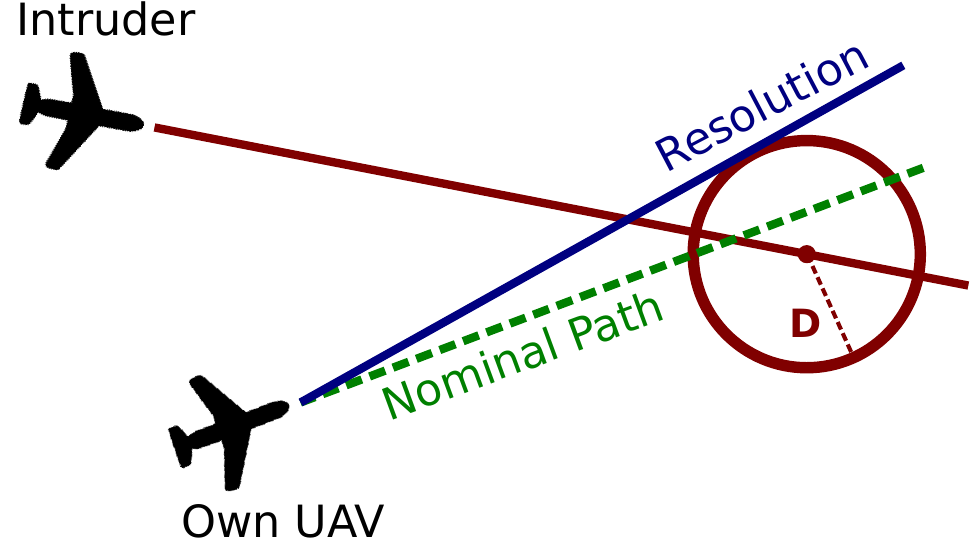}
    \caption{Trusted Resolution Logic. The near mid air collision exclusion zone (circle with radius $D$) moves with the intruder through time, but is shown here only at the time of closest approach. The TRL (\Cref{alg:trl}) finds a straight trajectory close to the nominal path that avoids this zone.}
    \captioncheat
    \label{fig:trl}
\end{figure}

We test the new approach in a scenario containing a UAV equipped with a perfect (noiseless) sensor to detect the state of an intruder and a simple TRL to resolve conflicts. This TRL, illustrated in \Cref{fig:trl}, determines a path that does not pass within a specified separation distance, $D$, of the intruder given that the intruder maintains its current heading (except in cases where no such path exists or when the TRL's heading resolution is too coarse to find such a path). To account for uncertainty in the intruder's flight path, if $D$ is fixed at a constant value, say $\bar{D}$, the value must be very large to ensure safety, but this may cause unnecessary departures from normal operation. To overcome this limitation, we compute an optimized policy $\tilde{\pi}$ that specifies a time varying separation distance, $D_t$, based on the encounter state. The goal is to ensure safety without being too conservative. 

The problem of dynamically selecting $D_t$ can be formulated as a Markov decision process (MDP). Online solution of $\tilde{\pi}$ using an algorithm such as Monte Carlo Tree Search (MCTS) \cite{AC-JH-NS-OT-NB:11} would be conceptually straightforward, but would require significant computing power onboard the vehicle. In addition, it would be difficult and time-consuming to rigorously certify the implementation of MCTS due to its reliance on pseudo-random number generation. The key contribution of this paper is to devise an \emph{offline} approach to compute $\tilde{\pi}$. Offline optimization of the policy is difficult because of the size of the state space of the MDP, which is the Cartesian product of the continuous state spaces of the UAV and the intruder. To overcome this challenge, we devise a value function approximation scheme that uses grid-based features that exploit the structure of the state space. This value function is optimized using simulation-based approximate dynamic programming (ADP). The policy is then encoded using an approximate post-decision state value function. With this post-decision value function, the UAV can easily extract the optimal action for the current state online by evaluating each possible action.

The remainder of the paper is organized as follows: \Cref{sec:enc} describes the MDP model  of an encounter between two UAVs, \Cref{sec:approach} describes a solution approach based on approximate dynamic programming, and \Cref{sec:results} contains numerical results followed by conclusions in \Cref{sec:conclusion}.

\section{Problem Formulation} \label{sec:enc}

% As discussed in the introduction, 
The problem of avoiding an intruder using an optimized TRL is formulated as an MDP, referred to as the \emph{encounter} MDP. An encounter involves two aerial vehicles flying in proximity. The first is the vehicle for which we are designing the resolution logic, which will be referred as the ``own UAV" (or simply ``UAV''). The second is the intruder vehicle, which may be manned or unmanned and will be referred to as the ``intruder." Throughout the paper, the superscripts $\own{}$ and $\intr{}$ refer to the own UAV and intruder quantities, respectively. Specifically, the encounter MDP is defined by the tuple $(S, A, T, R)$, which consists of

\begin{itemize}
    \item The state space, $S$: The state of the encounter, $s$, consists of (1) the state of the own UAV, $\rs$, (2) the state of the intruder, $\is$, and (3) a boolean variable $\dev$. The variable $\dev$ is set to true if the UAV has deviated from its nominal course and is included so that deviations at any point in time can be penalized equally. Collectively, the state is given by the triple
        \begin{equation}
            s = \left(\rs, \is, \dev\right) \text{.}
        \end{equation}
        The state components $\rs$ and $\is$ are specified in Section \ref{sec:veh}.
        To model termination, $S$ also includes a termination state, denoted $\sterm$.
    \item The action space, $A$: The actions are the possible values for the separation distance $D\in \reals_{\geq 0}$ used in the baseline TRL (see Figure \ref{fig:trl}).
    \item The state transition probability density function, $T: S \times A \times S \to \mathbb{R}$: The value $T(s,D,s')$ is the probability density of transitioning to state $s'$ given that the separation parameter $D$ is used within the TRL at state $s$. This function is implicitly defined by a generative model that consists of a state transition function $F(\cdot)$ (described in \Cref{sec:control}) and a stochastic process $W$ (described in \Cref{sec:veh}).
    \item The reward, $R:S \times A \to \mathbb{R}$: The reward function, defined in Section \ref{subsec:rew}, rewards reaching a goal and penalizes near mid air collisions (NMACs), deviation from the nominal path, and time outside a goal region.
\end{itemize}

\subsection{Model assumptions}\label{sec:assumptions}

Two important simplifying assumptions were made for this initial study. First, the UAV and the intruder move only in the horizontal plane and at constant speed. Modeling horizontal maneuvers is necessary because UAVs will likely have to employ them in place of or in addition to the vertical maneuvers that current collision avoidance systems for manned aircraft such as TCAS rely on. This is due to both climb performance limitations and potential regulatory constraints such as the \num{500}\si{ft} ceiling for small (\num{<55}\si{lb}) UAVs in proposed Federal Aviation Administration rules \cite{FAA:15}. Constraining altitude and speed simplifies exposition and reduces the size of the state space and hence the computational burden. Extensions to higher fidelity models (e.g.,~\cite{MJK-MWME-LPE-JKK-JDG:10}) are possible and left for future research. A higher fidelity model would present a computational challenge, but perhaps not an insurmountable one. For example, the TRL could be extended to handle variable speed and altitude, but the policy that governs the TRL parameters could be optimized only on the most important dimensions of the model (e.g., the horizontal plane). See~\cite{MJK-JPC-PPR:10} for a similar successful example.

Second, the intruder dynamics are \emph{independent} of the UAV's state; in other words, the intruder does not react to the flight path of the UAV. The cooperative setting where both the UAV and the intruder are equipped with a collision avoidance system (CAS) \cite{EJR:14,CT-GJP-SSS:98,MJK-JPC:11} is left for future research.

\subsection{Vehicle states and dynamics}\label{sec:veh}

This paper uses a very simple discrete-time model of an encounter between two aerial vehicles with time steps of duration $\Delta t$. Throughout this section, the superscript $\either{}$ may be replaced by either $\own{}$ or $\intr{}$. Both the UAV's and the intruder's states consist of the horizontal position $(x,y)$ and heading $\psi$, that is
\begin{equation}
    \rs = \left(\own{x}, \own{y}, \own{\psi} \right)\text{,} \quad 
    \is = \left(\intr{x}, \intr{y}, \intr{\psi} \right)\text{.}
\end{equation}
The UAV and intruder also have similar dynamics. Both aircraft fly forward in the horizontal plane at constant speeds denoted by $\either{v}$. They may turn at rates $\either{\dot{\psi}}$ that remain constant over the simulation step. The following equations define the vehicle dynamics: 
\begin{align*}
    \either{x}_{t+1} &= \begin{cases}
        \either{x}_{t} + \either{v} \cos\left(\either{\psi}_t\right)\Delta t & \text{if } \either{\dot{\psi}} = 0 \\
        \either{x}_{t} + \either{v} \frac{\sin\left( \either{\psi}_t + w_t\Delta t \right) - \sin\left( \either{\psi}_t \right)}{\either{\dot{\psi}}} & \text{otherwise}\\
    \end{cases} \\
    \either{y}_{t+1} &= \begin{cases}
        \either{y}_{t} + \either{v} \sin\left(\either{\psi}_t\right)\Delta t & \text{if } \either{\dot{\psi}} = 0 \\
        \either{y}_{t} - \either{v} \frac{\cos\left(\either{\psi}_t + \either{\dot{\psi}}\Delta t\right) - \cos\left(\either{\psi}_t\right)}{\either{\dot{\psi}}} & \text{otherwise}\\
    \end{cases} \\
    \either{\psi}_{t+1} &= \either{\psi}_t + \either{\dot{\psi}} \Delta t \text{.}
\end{align*}

The intruder makes small random turns with
\begin{equation}
    \intr{\dot{\psi}} = w_t \text{,}
\end{equation}
where $w_t$ is a stochastic disturbance. We let $W$ denote the stochastic process $\{w_t: t\in \naturals\}$. The random variables $w_t$ in $W$ are assumed independent and identically normally distributed with zero mean and a specified standard deviation, $\sigma_{\dot{\psi}}$. Subsequently, the intruder dynamics will be collectively referred to  as $\intr{f}\left(\intr{s}_t, w_t\right)$.

The dynamics of the  UAV are simplified conventional fixed wing aircraft dynamics with a single input, namely the roll angle $\own{\phi}$. We assume that the roll dynamics are  fast compared to the other system dynamics, so that the roll angle $\own{\phi}$  may be directly and instantaneously commanded by the control system. This assumption avoids the inclusion of roll dynamics, which would increase the size of the state space.
\begin{equation}
    \own{\dot{\psi}_t} = \frac{g \tan \own{\phi}_t}{\own{v}} \text{,}
\end{equation}
where $g$ is acceleration due to gravity. The performance of the UAV is limited by a maximum bank angle
\begin{eqnarray}
    |\own{\phi}_t| \leq \phi_\text{max} \text{.}
\end{eqnarray}
The UAV dynamics will be collectively referred to as $\own{f}\left(\own{s}_t, \own{\phi}_t\right)$.

\subsection{UAV control system and transition function} \label{sec:control}

The control system for the UAV consists of the TRL and a simple controller that commands a turn rate to track the course determined by the TRL. The TRL determines a heading angle, $\psires$, that is (1) close to the heading to the goal, $\psigoal$, and (2) will avoid future conflicts with the intruder given that the intruder maintains its current heading as shown in Figure~\ref{fig:trl}. In order to choose a suitable heading, many candidate headings, denoted $\psicand$, are evaluated. 

Given an initial state $s$, a candidate heading, $\psicand$, for the UAV, and that both vehicles maintain their heading, the distance between the vehicles at the time of closest approach is a simple analytical function. Specifically, consider the distance  $d\left(s,\psicand,\tau\right)$ between the vehicles $\tau$ time units in the future, i.e., 
\begin{equation} \label{eqn:dist}
    % d(s_t,\tau) = \sqrt{(\tau*\dot{x}^{(i)} + x^{(i)} - \tau*v^{(o)}*\cos(psi^{(o)}))^2 + (\tau*\dot{y}^{(i)} + y^{(i)} - \tau*v^{(o)}*\sin(\psi^{(o)}))^2}
    d\left(s, \psicand, \tau\right) = \sqrt{\Delta x(\tau)^2 + \Delta y(\tau)^2} \text{,}
\end{equation}
where
\begin{align*}
    \Delta x(\tau) &= \intr{x} - \own{x} + \tau \intr{v} \cos\left(\intr{\psi}\right) - \tau \own{v} \cos\left(\psicand\right),\\
    \Delta y(\tau) &= \intr{y} - \own{y} + \tau \intr{v} \sin\left(\intr{\psi}\right) - \tau \own{v} \sin\left(\psicand\right) \text{.}
\end{align*}

The minimum distance between the two vehicles is analytically found by setting the time derivative of $d\left(s,\psicand, \tau\right)$ to zero. Specifically, the time at which the vehicles are closest is given by
\begin{equation}
    \tau_\text{min}\left(s,\psicand\right) = \max \left\{\frac{a + b}{c-2d}, 0\right\} \text{,}
\end{equation}
where
% since these are not super important, I left some kind of ugly formatting (i.e. no \left(, \right)) to keep them shorter
\begin{align*}
    a &:= -\intr{v}\intr{x} \cos(\intr{\psi}) - \intr{v} \intr{y} \sin(\intr{\psi}) \\
    b &:= \own{v} \intr{x} \cos(\psicand) + \own{v} \intr{y} \sin(\psicand) \\
    c &:= v^{(\ownid)\,2} + v^{(\intrid)\,2}\cos^2(\intr{\psi}) + v^{(\intrid)\,2}\sin^2(\intr{\psi}) \\
    d &:= \own{v}\intr{v} ( \cos(\intr{\psi})\cos(\psicand) + \sin(\intr{\psi})\sin(\psicand) ) \text{.}
\end{align*}
The minimum separation distance over all future time is then
\begin{equation}
    d_\text{min}\left(s, \psicand\right) := d\left(s, \psicand, \tau_\text{min}\left(s,\psicand\right)\right) \text{.}
\end{equation}

The TRL begins with a discrete set of potential heading values for the UAV. It then determines, for a desired separation distance $D$,  which of those will not result in a collision given that the UAV and  intruder maintain their headings. Finally, it selects the value from that set which is closest to $\psigoal$. The TRL is outlined in Algorithm~\ref{alg:trl}.

\renewcommand{\algorithmicrequire}{\textbf{Input:}}
\renewcommand{\algorithmicensure}{\textbf{Output:}}
\begin{algorithm}[tb]
    \caption{Trusted Resolution Logic}\label{alg:trl}
\begin{algorithmic}
    \Require Encounter state $s$, desired separation distance $D$
    \Ensure Resolution heading angle $\psires$
        \Function{$\text{TRL}$}{$s$,$D$}
        % \State $\Psi \gets \left\{\psicand-\pi:\pi/20:\psicand+\pi\right\}$  \Comment{range of values for heading}
        % \State $\Psi \gets \left\{\own{\psi} + n\pi/N : n \in \{-N,\ldots,0,\ldots,N\}, N>0. \right\}$ \\
        \State $\Psi \gets \left\{\own{\psi} + n\pi/N : n \in \{-N,\ldots,N\}, N>0 \right\}$
        % \Comment{range of values for heading}
        \State $D^* = \max_{\psicand \in \Psi}\, d_\text{min}\left(s, \psicand\right)$
        %\State $mindists \gets \left\{d_\text{min}(s, \psi) : \psi \in \Psi\right\}$
        %\State $maxmindist \gets \max(mindists)$
        \If{$D^*< D$} \Comment{conflict inescapable}
            \State $\Psi \gets \left\{\psicand \in \Psi : d_\text{min}\left(s, \psicand\right) = D^*\right\}$
            \State \Return $\argmin{\psicand \in \Psi}\,|\psicand-\psigoal|$
        \Else
            \State $\Psi \gets \left\{\psicand \in \Psi : d_\text{min}\left(s, \psicand\right) \geq D\right\}$
            \State \Return $\argmin{\psicand \in \Psi}\,|\psicand-\psigoal|$
        \EndIf
    \EndFunction
\end{algorithmic}
\end{algorithm}

Once the TRL has returned the desired heading, $\psires$, a low-level  controller determines the control input to the vehicle. We write this as
\begin{equation}
    \own{\dot{\psi}}_t = c\left(\own{s}_t, \psires \right) \text{,}
\end{equation}
where $c(\cdot)$ represents the low level controller.

The closed-looped dynamics for the state variables of the vehicles are then given by
\begin{align}
    \label{eq:idyn}
    \intr{s}_{t+1} &= \intr{f}\left(\intr{s}_t, w_t\right) \\
    \label{eq:odyn}
    \own{s_{t+1}} &= \own{f}\left(\own{s_t}, c\left(\own{s_t}, \text{TRL}(s_t, D_t)\right)\right) \text{.}
\end{align}
The UAV and the intruder dynamics are coupled \emph{only} through the TRL.

The goal region that the UAV is trying to reach is denoted by $S_\text{goal}$. This is the set of all states in $S$ for which $\left\|\left(\own{x}, \own{y}\right) - \left(\own{x}_{\text{goal}}, \own{y}_{\text{goal}}\right)\right\| \leq \dgoal$, where $\dgoal>0$ is a specified goal region radius, and $\left(\own{x}_{\text{goal}}, \own{y}_{\text{goal}}\right)$ is the goal center location.
A near mid-air collision (NMAC) occurs at time $t$ if the UAV and intruder are within a minimum separation distance, $\dnmac > 0$, that is if $\left\|\left(\own{x}_t, \own{y}_t\right) - \left(\intr{x}_t, \intr{y}_t\right)\right\| \leq \dnmac$.
If the UAV reaches the goal region at some time $t$, i.e., $s_t \in S_\text{goal}$, or if an NMAC occurs, the overall encounter state $s$  transitions to the terminal state $\sterm$ and remains there.  If the UAV performs a turn, $\dev$ is set to true because the vehicle has now deviated from the nominal straight path to the goal.

Let the state transition function, defined by \eqref{eq:idyn}, \eqref{eq:odyn}, and the special cases above, be denoted concisely as $F$ so that  
\begin{equation}\label{eq:sys_dyn}
    s_{t+1} = F(s_t, D_t, w_t) \text{,}
\end{equation}
where, as stated above, $D_t$ is the input to the system (the action in the MDP formulation).

\subsection{Reward}\label{subsec:rew}

In this paper, we minimize two competing metrics. The first is the risk of an NMAC. As in previous studies (e.g., \cite{JEH-MJK-WAO:13}), this aspect of performance is quantified using the \emph{risk ratio}, the number of NMACs with the control system divided by the number of NMACs without the control system. The second metric is the probability of any deviation from the nominal path. This metric was chosen (as opposed to a metric that penalizes the magnitude of the deviation) because any deviation from the normal operating plan might have a large cost in the form of disrupting schedules, preventing a mission from being completed, or requiring manual human monitoring. The MDP reward function is designed to encourage a policy that performs well with respect to these goals.  

Specifically, the total reward associated with an encounter is the sum of the stage-wise rewards throughout  the entire encounter
\begin{equation}\label{eqn:encrew}
    \sum_{t=0}^\infty R(s_t, D_t) \text{.}
\end{equation}

In order to meet both goals, the stage-wise reward is
\begin{align}\label{eqn:rew}
    % r(s,a) = - 1 + 100 \times \mathbf{1}\left\{s^{(o)} \in S^{(o)}_\text{goal}\right\} - \lambda \, \mathbf{1}\left\{d(s) < D_\text{NMAC}\right\}
    R(s_t,D_t) := & - c_\text{step} + r_\text{goal} \times \operatorname{in\_goal}\left(\own{s}_t\right) \nonumber\\
             & - c_\text{dev} \times \operatorname{deviates}(s_t,D_t) \nonumber\\
             & - \lambda \times \operatorname{is\_NMAC}(s_t) \text{,}
\end{align}
for positive constants $c_\text{step}$, $r_\text{goal}$, $c_\text{dev}$,  and $ \lambda$. The first term is a small constant cost accumulated in each step to push the policy to quickly reach the goal. The function $\operatorname{in\_goal}$ indicates that the UAV is within the goal region, so the second term is a reward for reaching the goal. The third term is a penalty for deviating from the nominal path. The function $\operatorname{deviates}$ returns $1$ if the action will cause a deviation from the nominal course and $0$ otherwise. It will only return $1$ if the vehicle has not previously deviated and $\dev$ is false, so the penalty may only occur once during an encounter. Constants $c_\text{step}$, $r_\text{goal}$, and $c_\text{dev} $ represent relative weightings for the terms that incentivize a policy that reaches the goal quickly and minimizes the probability of deviation. Example values for these constants  are given in \Cref{sec:results}. The fourth term is the cost for a collision. The weight $\lambda$ balances the two performance goals. We heuristically expect there to be a value of $\lambda$ for which the solution to the MDP meets the desired risk ratio if it is attainable. Bisection, or even a simple sweep of values can be used to find a suitable value, and this method has been used previously to analyze the performance of aircraft collision avoidance systems \cite{HB-DH-MJK-WSL:12,MJK-JPC:11}.

\subsection{Problem statement}
The problem we consider  is to find a feedback control policy $\pi^*: S \to A$, mapping an encounter state $s_t$ into a separation distance $D_t$,  that maximizes the expected reward \eqref{eqn:encrew} subject to the system dynamics \eqref{eq:sys_dyn}:
\begin{equation}\label{eq:prob}
\begin{aligned}
& \underset{\pi}{\text{maximize}}
& & E \left[\sum_{t=0}^\infty R(s_t, \pi(s_t))\right] \\
& \text{subject to}
& & s_{t+1} = F(s_t, \pi(s_t), w_t) \text{,}
\end{aligned}
\end{equation}
for all initial states $s_0 \in S$. In the next section, we present a practical approach to problem \eqref{eq:prob} that uses approximate dynamic programming to find a suboptimal policy $\tilde{\pi}$.

\section{Solution Approach} \label{sec:approach}

The solution approach for problem \eqref{eq:prob} is an approximate dynamic programming algorithm called approximate value iteration \cite{DB:05}. The value function, $V$, represents the expected value of the future reward given that the encounter is in state $s$ and an optimal policy will be executed in the future. We approximate $V$ with a linear architecture of the form
\begin{equation}\label{eqn:val}
    \tilde{V}(s) = \beta(s)^\top \theta \text{,}
\end{equation}
where the feature function $\beta$ returns a vector of $N_\beta$ feature values, and $\theta\in \reals^{N_\beta}$ is a vector of weights \cite{DB:05}. At each step of  value iteration, the weight vector $\theta$ is fitted to the results of a large number of single-step simulations by solving a linear least-squares problem. After value iteration has converged, $\tilde{V}$ is used to compute a linear approximation of the \emph{post-decision state} value function, $\tilde{V}_q$. The policy is extracted online in real time by selecting the action that results in the post-decision state that has the highest value according to $\tilde{V}_q$. The choice of working with post-decision states will be discussed in \Cref{sec:extract}.

\subsection{Approximate value iteration} \label{sec:iter}

The bulk of the computation is carried out offline before vehicle deployment using simulation. Specifically, the first step is to estimate the optimal value function for problem \eqref{eq:prob} using value iteration \cite{DB:05}. On a continuous state space, the Bellman operator used in value iteration cannot be applied for each of the uncountably infinite number of states, so an approximation must be used. In this paper we adopt projected value iteration \cite{DB:05}, which  uses a finite number of parameters to approximate the value function. Each successive approximation, $\tilde{V}_k$, is the result of the Bellman operation projected onto a linear subspace, that is
\begin{equation}\label{eqn:projvi}
    \tilde{V}_{k+1}(s) = \Pi \mathcal{B}[\tilde{V}_k](s) \text{,}
\end{equation}
where $\mathcal{B}$ is the Bellman operator, and $\Pi$ is a Euclidean projection onto the linear subspace $\Phi$ spanned by the $N_\beta$ basis functions (see \cite{DB:05} for a detailed discussion of this approach).

To perform the approximate value iteration \eqref{eqn:projvi}, we resort to Monte Carlo simulations. Specifically, for each iteration, $N_\text{state}$ states are uniformly randomly selected. If the states lie within the grids used in the feature function (see \Cref{sec:features}) the sample is ``snapped'' to the nearest grid point to prevent approximation errors due to the Gibbs phenomenon \cite{JF-FBR:91}. At each sampled state $s^{[n]}$, $n=1,\ldots, N_\text{state}$, the stage-wise reward and the expectation of the value function are evaluated for each action $a$ within a discrete approximation of the action space $A$, denoted by $\tilde A$. The expectation embedded in the Bellman operator is approximated using $N_{\text{EV}}$ \emph{single-step} intruder simulations, each with a randomly generated noise value, $w_m$, $m=1,\ldots, N_{\text{EV}}$. However, since the UAV dynamics are deterministic, only one UAV simulation is needed. The maximum over $\tilde{A}$ is stored as the $n$th entry of a vector $v_{k+1}$:
\begin{multline}
    v_{k+1}[n] := \\
    \max_{D \in \tilde{A}} \left\{R(s^{[n]},D) + \frac{1}{N_{\text{EV}}}\sum_{m=1}^{N_{\text{EV}}} \beta(F(s^{[n]}, D, w_m))^\top \theta_k \right\} \text{,} \nonumber
\end{multline}
for $n=1,\ldots, N_\text{state}$. Here $v_{k+1}$ provides an approximation to the (unprojected) value function. To project $v_{k+1}$ onto $\Phi$, we compute the weight vector $\theta_{k+1}$ by solving the least-squares optimization problem
\begin{equation}
    \theta_{k+1} = \argmin{\theta\in \reals^{N_\beta}} \sum_{n=1}^{N_\text{state}} \left( \beta\left(s^{[n]}\right)^\top \theta - v_{k+1}[n] \right) ^2 \text{.}
\end{equation}
Iteration is terminated after a fixed number of steps, $N_{VI}$, and the resulting weight vector, denoted $\theta$, is stored for the next processing step (Section \ref{sec:extract}).

\subsection{Post decision value function extraction} \label{sec:extract}

For reasons discussed in \Cref{sec:policy}, our second step is to approximate a value function, $V_q$, defined over post-decision states \cite{DB:05,RSS-AGB:98}. A post-decision state, $q$, is a state in $S$ made up of the own UAV state and $\dev$ \emph{at one time step into the future} and the intruder state \emph{at the current time}, that is
\begin{equation} \label{eqn:pd}
    q_t := \left(\rs_{t+1}, \is_t, \dev_{t+1}\right) \text{.}
\end{equation}
Correspondingly, let $g:S\times A\to S$ be the function that maps the current state and action to the post-decision state. In other words, function $g(s_t,D_t)$ returns $q_t$ consisting of
\begin{align}\label{eqn:g}
    \rs_{t+1} &= \own{f}\left(\rs_t, c\left(\rs_t, TRL(s_t, D_t)\right)\right) \nonumber\\
    \is_t     &= \is_t \nonumber\\
    \dev_{t+1} &= \max\{\dev, \text{deviates}(s_t,D_t)\} \text{.}
\end{align}

The post-decision value function approximation, $\tilde{V}_q$, is computed as follows: Let $h:S \times \reals \to S$ be a function that returns the next encounter state given the post decision state and intruder heading noise value, that is $h(q_t,w)$ returns $s_{t+1}$ consisting of
\begin{align}
    \own{s}_{t+1} &= \own{s}_{t+1} \\
    \intr{s}_{t+1} &= \intr{f}(\intr{s}_{t}, w) \\
    \dev_{t+1} &= \dev_{t+1} \text{,}
\end{align}
where $\left(\own{s}_{t+1}, \intr{s}_t, \dev_{t+1}\right)$ are the components of $q_t$.

The post-decision value function, $V_q$, is, in terms of the value function, $V$,
\begin{equation} \label{eqn:pdexp}
    % V_q(q) = \underset{s'}{E} \left[ V(s') | q \right] ,
    V_q(q) = E \left[ V \left(h\left(q,w\right)\right) \right] \text{,}
    % not sure whether this notation is correct at all
\end{equation}
where $w$ denotes, as usual, a random variable with Gaussian normal density. The approximation, $\tilde{V}_q$, is of the linear form
\begin{equation}
    \tilde{V}_q(q) = \beta(q)^\top \theta^q \text{,}
\end{equation}
where $\beta(q)$ is the  feature vector for post-decision states $q\in S$, and $\theta^q \in \reals^{N_\beta}$ is the corresponding weight vector.

To calculate the weight vector, $N_q$ post decision states are randomly selected using the same method as described in \Cref{sec:iter} and are denoted $q^{[n]}$, $n=1,\ldots, N_q$. For each sample $q^{[n]}$, the expectation in (\ref{eqn:pdexp}) is approximated using $N_{\text{EV}}$ single-step simulations. The results are used to solve a least squares optimization problem
\begin{equation}
    \theta^q = \argmin{\theta \in \reals^{N_{\beta}}} \sum_{n=1}^{N_q} \left( \beta \left(q^{[n]}\right)^\top \theta - v^q[n]\right) ^2 \text{,}
\end{equation}
where
\begin{equation} \label{eq:pdvalue}
    v^q[n] := \frac{1}{N_{\text{EV}}} \sum_{m=1}^{N_{\text{EV}}} \beta\left(h\left(q^{[n]},w_m\right)\right)^\top \theta \text{,}
\end{equation}
where $w_m$, $m=1,\dots,N_\text{EV}$, is sampled from a random variable in $W$.

%The post decision value function is not used for the value iteration portion of the offline solution because it is less numerically convenient. As noted above, if the standard value function $\tilde{V}$ is used, the own UAV dynamics and TRL must be simulated only once to calculate the expectation in (\ref{eqn:bellman}). If, on the other hand, $\tilde{V}_q$ is used, the Bellman operator becomes
%    \begin{equation}
%        \mathcal{B}[\tilde{V}_q](q) = \underset{s}{E}\left[ \left. \max_{a \in \tilde{A}}\left\{ R(s,a) + \tilde{V}_q(g(s,a)) \right\} \,\right| q \right] .
%    \end{equation}
%    Since the own UAV dynamics and TRL calculations are part of $g$, they would have to be evaluated at every sample used to estimate the expectation.

\subsection{Online policy evaluation} \label{sec:policy}
The first two steps (explained, respectively, in Sections \ref{sec:iter} and \ref{sec:extract}) are performed offline. The last step, namely policy evaluation, is performed online. Specifically a suboptimal control at any state $s$ is computed  as
\begin{equation}\label{eq:post}
    \tilde{\pi}(s) = \argmax{D \in \tilde{A}} \, \tilde{V}_q\left(g(s,D)\right) \text{.}
\end{equation}
Since $g$ (defined in \eqref{eqn:g}) is a \emph{deterministic} function of a state-action pair, this calculation does not contain any computationally costly or difficult-to-certify operations such as estimating an expectation.

Two comments regarding the motivation for using the post-decision value function are in order. First, 
if the value functions could be exactly calculated, the post-decision state approach would be equivalent to the more common state-action value function approach, wherein a control is computed by solving
\begin{equation}
    \pi(s) = \argmax{D \in \tilde{A}} \, Q(s,D) \text{.}
\end{equation}
Here, $Q(s,a)$ is the state-action value function representing expected value of taking action $a$ in state $s$ and then following an optimal policy \cite{DB:05,RSS-AGB:98}. One can readily show that, in the exact case, $Q(s,D) = V_q\left( g(s,D) \right)$. However, for this problem, post-decision states are much less susceptible to approximation errors. This is primarily due to the fact that it is difficult to specify suitable features in the state-action domain. Since $g$ is highly nonlinear (indeed it is discontinuous because of the TRL), a grid interpolation approximation scheme (see \Cref{sec:features}) is not suitable for approximating $Q$. However, by evaluating $g$ online at the current state in \Cref{eq:post}, the difficulties with nonlinearity are avoided since $V_q$ is well-approximated by grid interpolation. 

Second, the post decision value function is \emph{not} used for the value iteration portion of the offline solution as the expectation estimate in \Cref{eq:pdvalue} would require $N_{EV}$ simulations of both the own UAV and the intruder dynamics and, therefore, would be more computationally demanding.

% when approximations are used, post-decision states provide a more robust (i.e., less susceptible to approximation error) way of deriving control actions \cite{DB:05}. One reason for this is that the cost function is approximated in the space of post-decision states, rather than in the larger space of state-control pairs, and hence the post-decision method is less susceptible to complications with inadequate exploration \cite{DB:05}. 

\subsection{Selection of features} \label{sec:features}

The primary value function approximation features are the interpolation weights for points in a grid \cite{SD:97}. A grid-based approximation is potentially inefficient compared to a small number of global features (e.g. heading, distance, and trigonometric functions of those variables). However, it is well known that the function approximation used in value iteration must have suitable convergence properties~\cite{JAB-AWM:95} in addition to approximating the final value function. Indeed, we experimented with a small number of global features, but were unable to achieve convergence and resorted to using a grid. Since a grid defined over the entire six dimensional encounter state space with a reasonable resolution would require far too many points to be computationally feasible, the grid must be focused on important parts of the state space.

Our strategy is to separate features into two groups (along with a constant), specifically
\begin{equation}
    \beta(s) = [\beta_\text{intruder}(\own{s}-\intr{s}), \beta_\text{goal}(\own{s}), 1] \text{.}
\end{equation}
The first group, $\beta_\text{intruder}$, captures the features corresponding to a near midair collision and is a function of only the position and orientation of the UAV relative to the intruder. The second group, $\beta_\text{goal}$, captures the value of being near the goal and is a function of only the UAV state.

\begin{figure}[tb]
    \centering
    \includegraphics[width=\columnwidth]{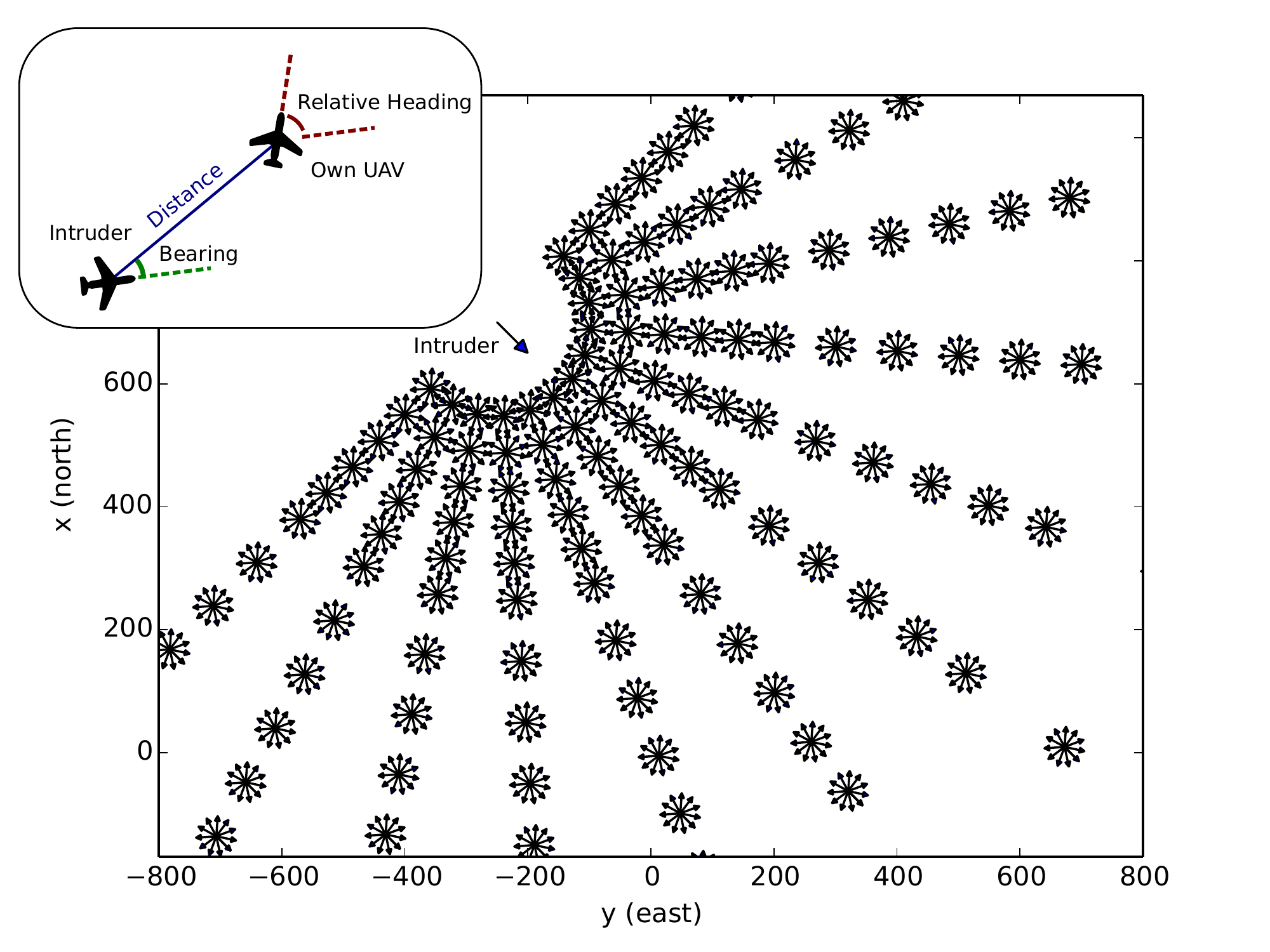}
    \caption{Intruder interpolation grid for $\beta_\text{intruder}$, visualized with the intruder at (\SI{700}{m}, \SI{-250}{m}) at heading \num{135}. The top left inset shows the variables used. In the main plot, each of the small arrows represents a grid point.}
%At each of the point locations, there are twelve small arrows radiating out. Each arrow represents a different own UAV heading.}
    \captioncheat
    \label{fig:intrudergrid}
\end{figure}

% \begin{figure}[tb]
%     \centering
%     \includegraphics[width=\columnwidth]{figures/goal_grid_plus.eps}
%     \caption{Goal interpolation grid for $\beta_\text{goal}$ when the own UAV's heading is directly north. The grid takes advantage of symmetry in the bearing variable, so, for each arrow on the left side, there is an arrow on the right side that corresponds with the \emph{same} point in the grid.}
%     \captioncheat
%     \label{fig:goalgrid}
% \end{figure}

Since the domain of $\beta_\text{intruder}$ is only three dimensional and the domain of $\beta_\text{goal}$ is only two dimensional, relatively fine interpolation grids can be used for value function approximation without requiring a prohibitively large number of features. The $\beta_\text{intruder}$ feature group consists of a NMAC indicator function and interpolation weights for a grid (\Cref{fig:intrudergrid}) with nodes at regularly spaced points along the following three variables: (1) the distance between the UAV and the intruder, (2) the bearing from the intruder to the UAV, and (3) the relative heading between the vehicles. The $\beta_\text{goal}$ vector consists of a goal indicator function, the distance between the UAV and the goal, and interpolation weights for a grid with nodes regularly spaced along the distance between the UAV and the goal and the absolute value of the bearing to the goal from the UAV. The total number of features is $N_\beta = 1813$.

\begin{figure}[tb]
    \centering
    \includegraphics[width=0.8\columnwidth]{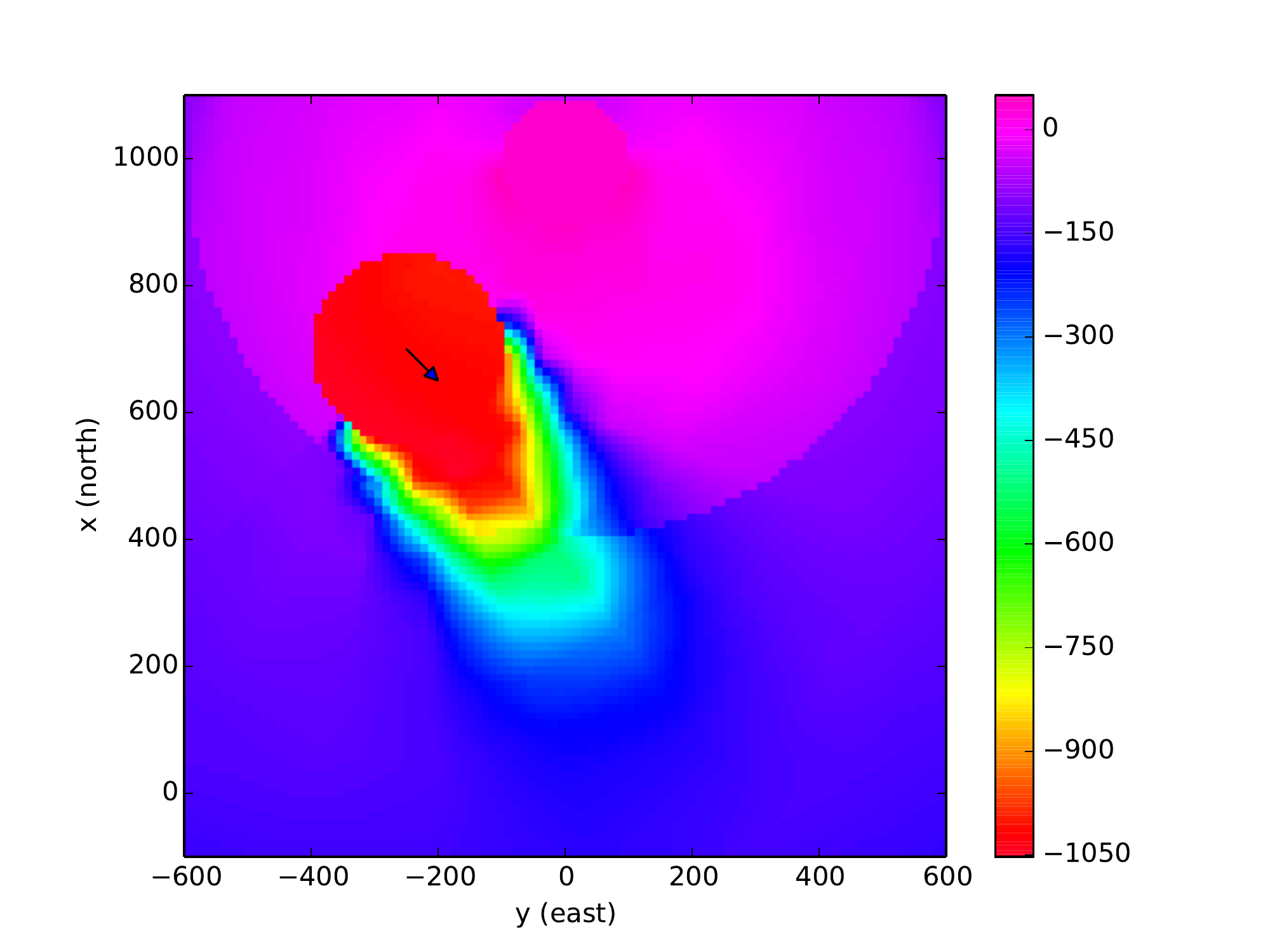}
    % \missingfigure{Value Function}
    \caption{Two-dimensional slice of the approximate optimal value function. Each pixel on this image represents the value function evaluated with the intruder at that position pointed directly north. The intruder is located at (\SI{700}{m},\SI{-250}{m}) pointed at heading 135 as indicated by the arrow. The goal is at (\SI{1000}{m},\num{0}).}
    \label{fig:val}
\end{figure}

Figure~\ref{fig:val} shows a single two-dimensional ``slice'' of the full six-dimensional optimized value function. As expected, there is a low value region in front of and to the south of the intruder and an increase in the value near the goal.

\begin{figure}[tb]
    \centering
    \includegraphics[width=0.8\columnwidth]{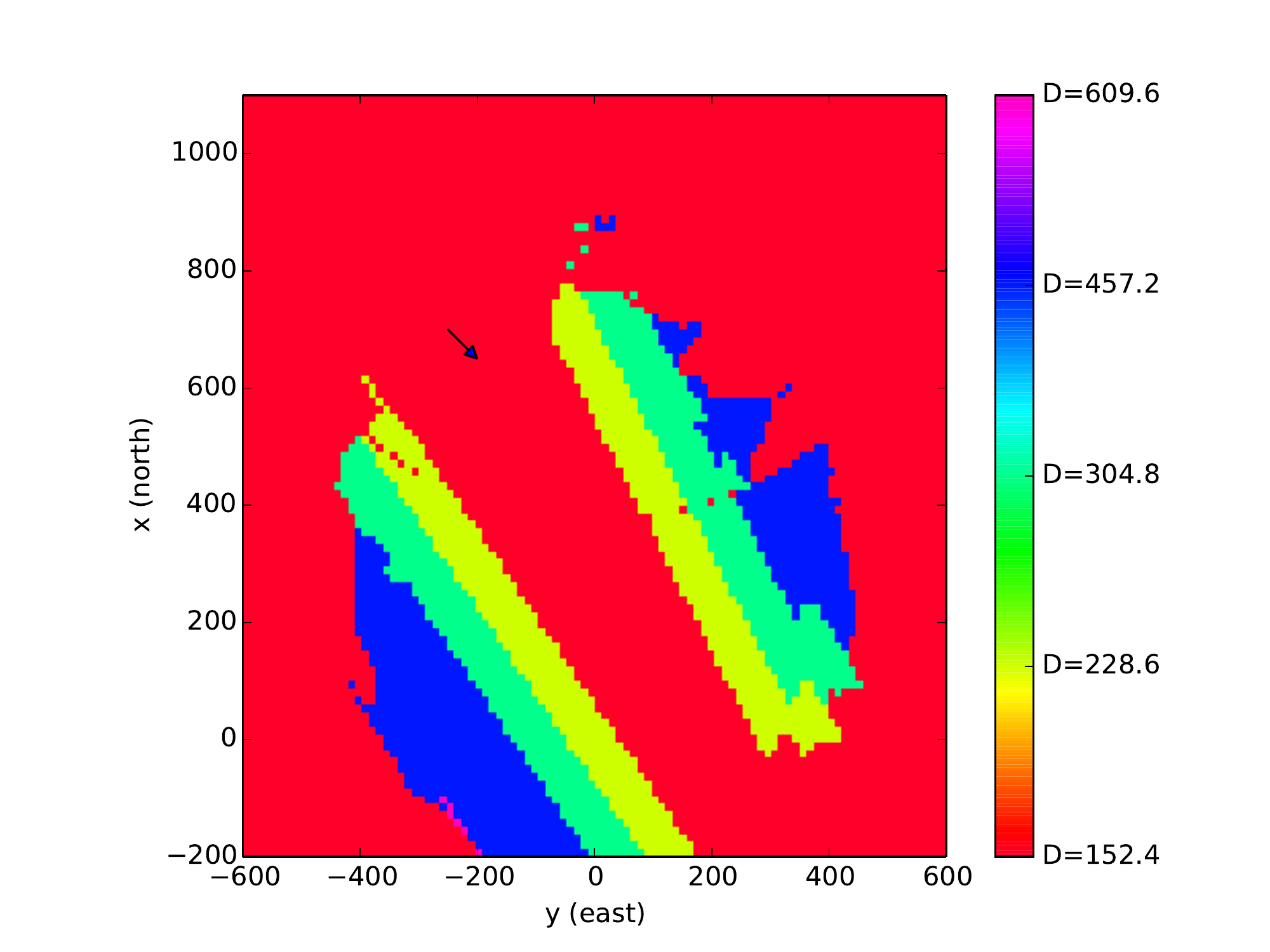}
    \caption{Two-dimensional slice of the optimized policy for TRL parameter $D$. Each pixel in the image is the policy evaluated with the own UAV at that location pointed directly north. The intruder is located at (\SI{700}{m},\SI{-250}{m}) and pointed at heading \num{135} as indicated by the arrow. Each color represents a different value for $D$ indicated by the number in the key on the right.}
    \captioncheat
    \label{fig:policy}
\end{figure}

\Cref{fig:policy} shows a slice of an optimized policy. When multiple actions result in the same post-decision state value, the least conservative action is chosen, so the policy yields the lowest value of $D$ ($\num{500} \si{ft} \approx \num{152.4} \si{m}$) on most of the state space. Because the own UAV is pointed north, the policy is conservative in a region in front of and to the south of the intruder. The band corresponding to small $D$ that stretches across the middle of the conservative region (from (\SI{700}{m},\SI{-250}{m}) to (\SI{-100}{m},\SI{200}{m})) is present because all values of $D$ result in the same post decision value.

\section{Results} \label{sec:results}

This section presents results from numerical experiments that illustrate the effectiveness of our approach. The experiments are designed to compare the three approaches discussed in \Cref{sec:intro}: the ``static TRL'' approach, the ``directly optimized'' approach, and the newly proposed ``optimized TRL'' approach. Each is represented by a different control law that specifies a turn rate to the own UAV. Specifically, the static TRL law uses the TRL described in \Cref{alg:trl} with a constant value for the separation distance (denoted $\bar{D}$). To represent the directly optimized approach, an optimized value function approximation and policy are generated using the method described in \Cref{sec:approach}, except that the policy \emph{directly determines the turn rate, $\own{\dot{\psi}}_t$, instead of specifying $D$}. The action space for this policy is $\{-\dot{\psi}_\text{max},-\dot{\psi}_\text{max}/2,0,\dot{\psi}_\text{max}/2,\dot{\psi}_\text{max}\}$. The third optimized TRL law works as described in \Cref{sec:enc,sec:approach}, dynamically assigning a value of $D$ to the TRL based on the state and using $c(\cdot)$ to assign the turn rate. The action space for the underlying $D$ policy is $\tilde{A}=\{\dnmac, 1.5 \dnmac, 2 \dnmac, 3 \dnmac, 4 \dnmac \}$.

The parameters used in the numerical trials are listed in \Cref{tab:params}. The control policies are evaluated by executing them in a large number of complete (from $t=0$ to the end state) encounter simulations. The same random numbers used to generate intruder noise were reused across all collision avoidance  strategies to ensure fairness of comparisons. In each of the simulations, the own UAV starts pointed north at position $(0,0)$ in a north-east coordinate system with the goal at $(1000\si{m},0)$.

The evaluation simulations use the same intruder random turn rate model with standard deviation $\sigma_{\dot{\psi}}$ that was used for value iteration. A robustness study using different models is not presented here, but previous research~\cite{MJK-JPC-PPR:10} suggests that this method will offer good performance when evaluated against both a range of noise parameters and structurally different models. The intruder initial position is randomly generated between \SI{800}{m} and \SI{1500}{m} from the center point of the encounter area at $(500\si{m},500\si{m})$ with an initial heading that is within \ang{135} of the direction from the initial position to the center point. 

The conservativeness of each control law is characterized by counting the number of deviations from the nominal path in \num{10000} simulations. Of these simulations, \num{1009} result in a NMAC if the UAV follows its nominal path, but for most a deviation would not be necessary to avoid the intruder.

The risk ratio is estimated using a separate set of \num{10000} simulations. Each of these simulations has an initial condition in the same region described above, but initial conditions and noise trajectories are chosen by filtering random trials so that each of the simulations \emph{will result in a NMAC if the own UAV follows its nominal path}. The risk ratio estimate for a policy is simply the fraction of these simulations that result in a NMAC when the policy is executed on them.
\begin{table}[tb]
    \caption{Parameters for numerical experiments} \label{tab:params}
    \centering
    % SHORTEN: make tiny
    % \tiny{
    \begin{tabular}{p{0.6\columnwidth} l r}
        \toprule
        Description & Symbol & Value \\
        \midrule
        Own UAV speed & $\own{v}$ & \num{30} \si{m/s} \\
        Maximum own UAV turn rate & $\dot{\psi}_\text{max}$ & \ang{18.7}\si{/s} \\
        Intruder speed & $\intr{v}$ & \num{60} \si{m/s} \\
        Intruder turn rate standard deviation & $\sigma_{\dot{\psi}}$ & \ang{10}\si{/s} \\
        Near mid air collision radius & $\dnmac$ & \num{500} \si{ft} \\
        Step cost & $c_\text{step}$ & \num{1} \\
        Reward for reaching goal & $r_\text{goal}$ & \num{100} \\
        Cost for deviation & $c_\text{dev}$ & \num{100} \\
        Step simulations for expectation estimate & $N_{EV}$ & \num{20} \\
        Single step simulations per round of value iteration (optimized TRL) & $N_\text{state}$ & \num{10000} \\
        Single step simulations per round of value iteration (directly optimized) & $N_\text{state}$ & \num{50000} \\
        Number of value iteration rounds & $N_{VI}$ & \num{35} \\
        Single step simulations for post decision value function extraction & $N_q$ & \num{50000} \\
        \bottomrule
    \end{tabular}
% }
\end{table}

% \begin{figure}[tb]
%     \centering
%     \includegraphics[width=0.8\columnwidth]{figures/intruderics.png}
%     \caption{Intruder initial conditions for evaluation simulations. Each small black arrow is an intruder initial state. The large green arrow is the controlled aircraft's initial state. The red dot is the goal.}
%     \label{fig:intruderics}
%     % I know this figure takes up a lot of room, but it is very difficult to describe in text
% \end{figure}

\begin{figure}[tb]
    \centering
    % \includegraphics[width=\columnwidth]{trl_paretto.pdf}
    % \missingfigure{Paretto curves}
    \includegraphics[width=0.8\columnwidth]{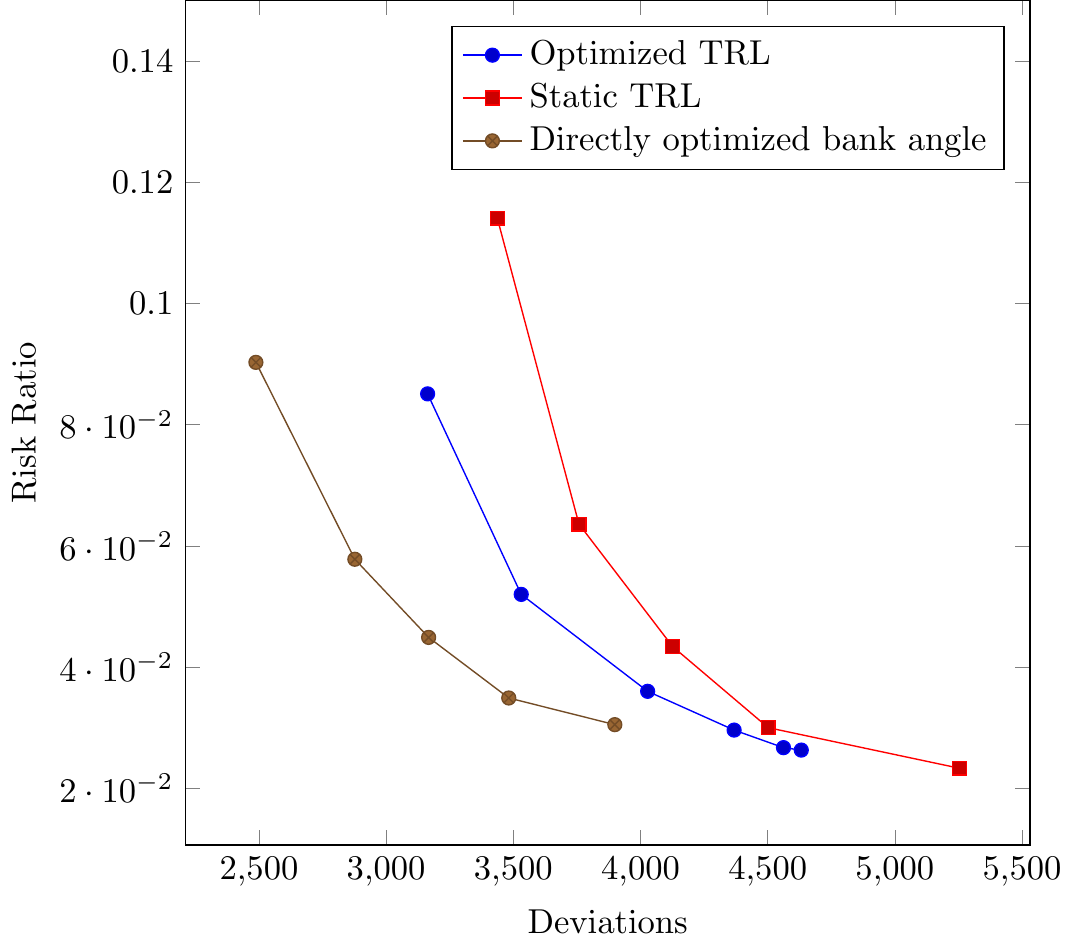}
    \caption{Control policy comparison. The horizontal axis variable is the number of deviations in \num{10000} encounter simulations. The values of $\lambda$ used to generate the datapoints are \num{100}, \num{316}, \num{1000}, \num{3160}, \num{e4}, and \num{3.16e4} for the optimized TRL policy and \num{300}, \num{500}, \num{700}, \num{1000}, and \num{1500} for the directly optimized approach. The values of $\bar{D}$ for the static TRL policy are \SI{250}{m}, \SI{300}{m}, \SI{350}{m}, \SI{400}{m}, and \SI{500}{m}.}
    \captioncheat
       \label{fig:pareto}
\end{figure}

\Cref{fig:pareto} shows the Pareto optimal frontiers for the different control laws. Each curve is generated by using various values of $\lambda$ in the reward function of the MDP, or by using various static values of $\bar{D}$ in the static TRL case. It is clear that optimization improves the performance of the TRL. For example, if the desired risk ratio is $5\%$, interpolation between data points suggests that the optimized TRL policy will cause approximately $10\%$ fewer deviations than the static TRL policy.

The directly optimized approach performs better than both TRL policies. This is not unexpected since the directly optimized policy is not limited to trajectories that the TRL deems to be safe. However, precisely because the directly optimized policy is not limited to trajectories guaranteed safe by hand-specified rules, it is much more difficult to convince people to trust it. These results thus allow us to estimate the performance \emph{price} of using a trusted resolution logic rather than an untrusted one. Since the optimized TRL causes approximately $18\%$ more deviations than the directly optimized policy at a risk ratio of \num{0.05}, we may say the price of the trustworthy properties of the TRL at a desired risk ratio of \num{0.05} is an $18\%$ increase in deviations. The new optimized TRL approach has reduced this from a price of a $31\%$ increase in deviations without optimization.

The Julia code used for these experiments is available at \texttt{\small{https://github.com/StanfordASL/UASEncounter}}.

\section{Conclusion} \label{sec:conclusion}

This paper presented a method for improving the performance of trusted conflict resolution logic for an unmanned aerial vehicle through approximate dynamic programming. A linear value function approximation based on interpolation grids and other features produces policies that improve the performance of the resolution logic without undermining the trust placed in it. Specifically, simulation experiments show that this optimization approach is able to decrease the number of deviations from the nominal path without increasing the risk ratio. Comparison with a directly optimized approach that is more difficult to trust can help to quantify the price of the gain in trustworthiness that comes from using the trusted resolution logic. The optimization approach proposed here is capable of reducing that price.

% There is much potential for future work on this problem. The current formulation models uncertainty in the intruder's actions, but sensor uncertainty should also be taken into account. If sensor uncertainty is added to the problem, it becomes a partially observable Markov decision process. As mentioned in the introduction, one promising solution approach for the partially observable version of this problem is MCVI \cite{HB-DH-MJK-WSL:12}.Also, as mentioned in \Cref{sec:assumptions}, there is potential to expand this work to more complicated scenarios with multiple intruders, intruders equipped with a CAS, higher fidelity models, or different sets of assumptions.

There are several directions for future work on this problem. The current formulation models uncertainty in the intruder's actions, but sensor uncertainty should also be taken into account. If sensor uncertainty is added to the problem, it becomes a partially observable Markov decision process. One promising solution approach for the partially observable version of this problem is Monte Carlo value iteration~\cite{HB-DH-MJK-WSL:12}. This approach has been applied to UAV collision avoidance before~\cite{HB-DH-MJK-WSL:12}, but it would be difficult to certify as a direct control system. Monte Carlo value iteration used in conjunction with a TRL has potential to handle both uncertainty in intruder behavior and sensor measurement noise while being easily certifiable. Also, as mentioned in \Cref{sec:assumptions}, there is potential to expand this work to more complicated scenarios with multiple intruders, intruders equipped with a CAS, higher fidelity models, or different sets of assumptions.

% There is also potential to expand this work to more complicated scenarios with multiple intruders and \mpmargin{cooperative vehicles}{vertical maneuvers? Make sure all assumptions are represented in this discussion}.
 
\subsubsection*{Acknowledgments}

This material is based on work supported in part by the National Science Foundation, Grant No. 1252521 and in part by the Office of Naval Research, Science of Autonomy Program, Contract N00014-15-1-2673.

\bibliographystyle{IEEEtran}
\bibliography{../../../bib/alias,../../../bib/main}

\end{document}